# Comparative Study of Instance Based Learning and Back Propagation for Classification Problems

[1]Nadia Kanwal, [2]Erkan Bostanci

[1]Department of Computer Science, Lahore College for Women University, Lahore, Pakistan

[2]Computer Engineering Department, Ankara University, Turkey

## Abstract

The paper presents a comparative study of the performance of Back Propagation and Instance Based Learning Algorithm for classification tasks. The study is carried out by a series of experiments will all possible combinations of parameter values for the algorithms under evaluation. The algorithm's classification accuracy is compared over a range of datasets and measurements like Cross Validation, Kappa Statistics, Root Mean Squared Value and True Positive vs False Positive rate have been used to evaluate their performance. Along with performance comparison, techniques of handling missing values have also been compared that include Mean or Mode replacement and Multiple Imputation. The results showed that parameter adjustment plays vital role in improving an algorithm's accuracy and therefore, Back Propagation has shown better results as compared to Instance Based Learning. Furthermore, the problem of missing values was better handled by Multiple imputation method, however, not suitable for less amount of data. Keywords: Classification, Instance Based Learning, Back Propagation, Imputation method

## 1. Introduction

For Machines to learn, classification is the fundamental requirement, therefore a lot of algorithms have been developed to classify the unseen data according to already stored examples. There is large variety of algorithms but no one can give the best accuracy over all kinds of data as their behaviour changes according to data, some perform well over Categorical data while other classify Real data accurately. It is, therefore, required to evaluate the performance of these algorithms based on multiple datasets. This study has been conducted to find the strengths and weaknesses of two well known classifiers over number of data sets.

Second section of the paper gives a brief preface to classification algorithms. The detail of datasets used has been provided in section 3 which is then followed by software description used to carry out the experiments under section 4. Section 5 details the experimental considerations like parameters which have constant values, testing methods etc. Section 6 and 7 provides experiments results and analysis of the said results related to the algorithms under reference. Introduction and evaluation of handling missing value techniques is given in section 8. The conclusion drawn form the study forms the last part of the paper.

## 2. The Classifiers

The classifiers selected are Back Propagation (BP) and Instance Based Learning (IBL). Both algorithms belong to the class of supervised learning that uses training data to classify the new or unseen instance. Both of these algorithms have their own advantages. Instance Based Learning is considered good due to its simplicity, less time consumption while maintaining good level of accuracy. On the other hand Back Propagation achieve promising results due to its weighted error feed backward approach. A brief introduction of these two algorithms is given hereunder.

Instance Based Learning (IBL)

It uses a pretty simple approach of storing the training data and classifying a new unseen data by calculating the distance of the most similar example already stored. The algorithm has different implementations like IBL and IBK (K-nearest neighbour algorithm). Where IBK algorithm uses voting method to decide the classification of new example and the number of votes is denoted by "k" value. The value of k needs to be odd to avoid tie situations. There is another variation in this algorithm to assign weights to instances according to their distance for better classification. This is called Distance-weighted Nearest Neighbour Algorithm. The weight can be either inverse of the distance or the complement of the distance.

Back Propagation

Back propagation is a technique that applies to Multilayer Perceptron Networks to adjust the weights of hidden layers to reduce the error. It is named "back propagation" because it propagates the weighted error backward to hidden layers to update weight. For this it uses generalised delta rule.

The convergence of Back propagation to local minima for error depends on the value of $\alpha$ (learning rate coefficient), too large $\alpha$ makes it difficult for network to find the gradient (narrow peak) while too small $\alpha$ usually increase the chance of getting trapped into local maxima. The problem is solved using another term called momentum ($\beta$), which reduces the chance of getting stuck in local peak as well as it accelerates the learning over smooth surface. By changing the values of $\alpha$ and $\beta$ the network can be tuned to perform better.

## 3. Data Sets

Six data sets have been selected from UCI machine learning repository. The description of these datasets is given in table 1. Datasets are grouped into two categories one with missing values and other without missing values. The purpose of this grouping is to evaluate the performance of algorithms as well as to find better solution to handle missing values problem to reduce the error rate and in turn increase the accuracy of the algorithm.

| Data Set | Instances | Attributes | Class | Missing Values |
|---|---|---|---|---|
| Abalone | 4177 | 8 | 29 | -Nil- |
| Iris | 150 | 4 | 3 | -Nil- |
| Glass Identification | 214 | 9 | 7 | -Nil- |
| Echocardiogram | 132 | 13 | 2 | Yes |
| Ozone Level detection | 2536 | 73 | 2 | Yes |
| Breast Cancer | 699 | 10 | 2 | Yes |

Table 1: Data Sets and Their Description

## 4. The Software

To carry out all the experiments WEKA Machine Learning tool (1) has been used. WEKA has variety of implementations of Machine Learning and Data Mining algorithms. For Back Propagation WEKA implementation of Multilayer Perceptron is used. For Instance based learning "K-Nearest Neighbour algorithm" is used as IBK in WEKA machine.

Since three methods of handling missing values have been explored (Ignore Instance, Mean/Mode value replacement and Multiple Imputation). The first method i.e. Ignore Missing value in Back-Propagation is already implemented in WEKA as default missing value handling technique, for second method again WEKA provides the functionality of calculating the mean and replacing it with missing value denoted by "?". Final and last method is Multiple Imputation[1] (MI). It is implemented by using freely available software called "NORM" (1). The description of Multiple Imputation Method is given in section 8.1.NORM[2] is the implementation of Multiple Imputation of multivariate continuous data under a normal model. For this study the software is used to impute the best value of the attribute calculated by using Monte Carlo method (1).

---

[1] Statistical Method to find best suited value to replace missing value.

[2] Software for Multiple imputation

It is important to mention here that IBK does not ignore the missing value by default rather it assigns maximum distance to the attribute with missing value in one or both of the instances being compared. Therefore, for IBK, this method is considered as default method.

## 5. Experimental considerations

The performance evaluation is carried out by applying both algorithms to all data sets initially by using default values of the parameters assigned by WEKA. Since the experiment time could be effected due to system load therefore every experiment is repeated number of times to achieve the best approximation. It is also important to mention that all of the experiments are carried out using the same processor to have fair comparison.

**K-Nearest Neighbour:**

K-Nearest Neighbour Algorithm (IBK) with different values of K has been tested for all datasets. The value of K always kept odd to have quick and fair voting decision and avoid ties. The approach is to keep increasing the value of K until there is no significant change hence values 1, 3, 5, 7, 9, 11 and 15 have been tested and it was observed that the performance did not change or rather degrades for higher values of K. Also larger values of K can bring biasness in the decision while smaller values of k will bring variance in classification. Therefore the approach is to find K that can reduce the probability of misclassification which means a value of K not too large and not too small. Other parameters that remain the same for all experiments are given in table: 2 hereunder.

Cross validate is kept false because 10 fold cross validation is used in the experiments instead of leave-one-out validation. No normalization is kept on to have attribute normalization.

| Debug | False |
|---|---|
| NoNormalization | False |
| Cross Validate | False |

**Table 2: IBK, Default parameters setting**

**Multi Layer Perceptron Model:**

While applying Multilayer Perceptron model multiple combinations of learning rate ($\alpha$), Momentum ($\beta$), and number of hidden layers have been tested. More than 500 experiments have been conducted to find the best combination of $\alpha$, $\beta$, and hidden layers to achieve the least RMSE and maximum accuracy while keeping other parameters constant which are mentioned in the table: 3.

| Decay | False |
|---|---|
| Nominaltobinaryfilter | True |
| Normalizednumericlass | True |
| Trainingtime | 500 |
| Validationtionthreshold | 20 |

Table 3: MLP, Default parameters setting

Along with these, 10 fold cross validation has been used for testing. Most of the time large number of hidden layers up to 20 were tested but if there wasn't any improvement in accuracy or there was considerable increase in MAE, adding more hidden layers was considered futile. Another drawback of adding more layers is that it needs more time to train the network.

# 6  Performance Evaluation over Datasets

## 6.1  Abalone

The data set contains 29 classes to predict. Classes have the values from 1 to 29. The performance of both algorithms over this dataset is not significant as shown in the in Fig:1 below. This is due to the fact that the dataset does not contain sufficient examples for all classes, for example for classes 1,2,24,25,26,27, and 29 there are only one or two instances present in the data. However in comparison, Multi Layer Perceptron algorithm has well classified the data and achieved an accuracy of 26.22%. To avoid over fitting in back-propagation number of hidden layers and learning rate is kept small.

| | |
|---|---|
| MLP:       α=0.3,  β=0.2 , Hidden Layer=3 | |
| IBK:             K=11 | |
| IBK(1-d):      K=11 | |
| IBK(1/d):      K=11 | |

Table 4: Parameter Values used

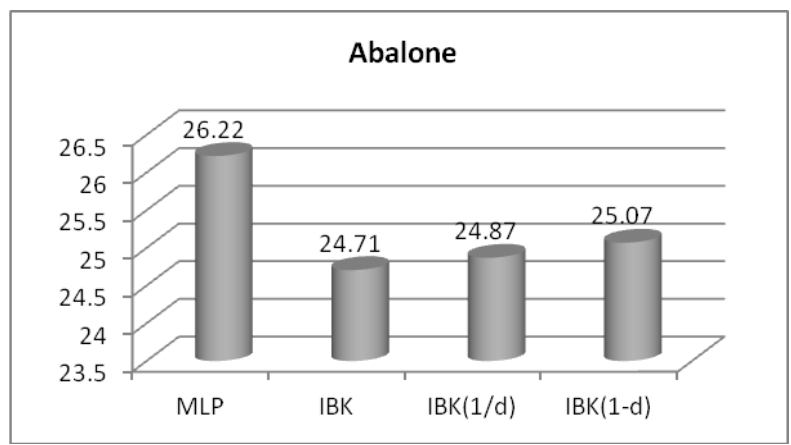

Fig. 1: Accuracy Comparison over Abalone

**Result:**  MLP is better than IBK

## 6.2 Echocardiogram

This dataset has 2 classes, but most of the attributes have missing data. MLP did not perform well over this dataset due to large number of missing values. The experiment was repeated upto 20 hidden layers but the best combination of accuracy and RMSE achieved for the following values learning rate, momentum and hidden layer. On the other hand IBK (1/d) classified very well using 11 neighbours to vote. The accuracy achieved by algorithms is shown in Fig.:2. Since the number of instances is low and 10 fold cross validation takes 10% of data randomly to test the model, hence increase in number of hidden units could cause over fitting and in turn can affect the accuracy.

| MLP: | $\alpha=0.2$ , $\beta=0.2$ ,Hidden Layer=1 |
|---|---|
| IBK: | K=11 |
| IBK(1-d): | K=11 |
| IBK(1/d): | K=11 |

Table 5: Parameter Values used for Echocardiogram

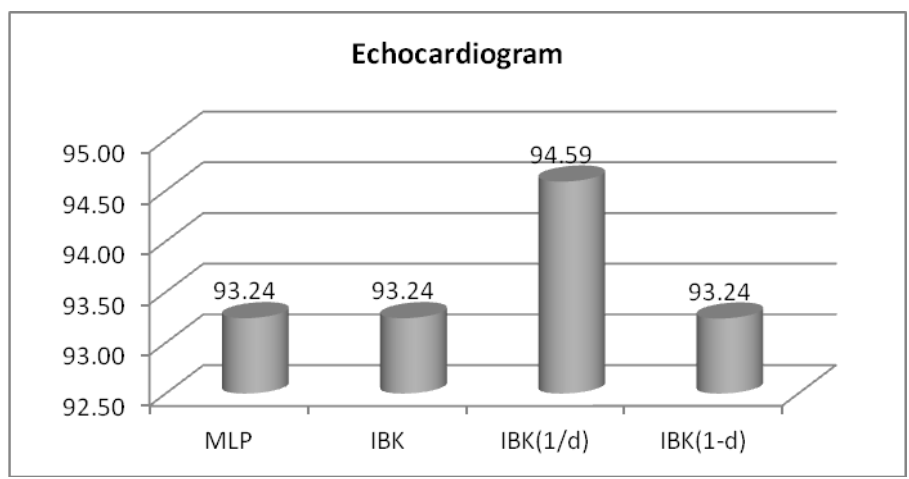

Fig. 2: Accuracy Comparison over Echocardiogram

**Result:** IBK (1/d) performed well over MLP

## 6.3 Glass Identification

The data set contains data about glass identification with information recorded in 9 attributes and by using this information the requirement is to classify the data for 7 classes. The behaviour of algorithms is apparent in the graph. The parameter values have been selected after series of experiments to attain a balance between good accuracy and reduced over fitting.

| | |
|---|---|
| MLP: | α=0.5, β=0.5 ,Hidden Layer=4 |
| IBK: | K=1 |
| IBK(1-d): | K=1 |
| IBK(1/d): | K=1 |

Table 6: Parameter Values used for Glass Identification

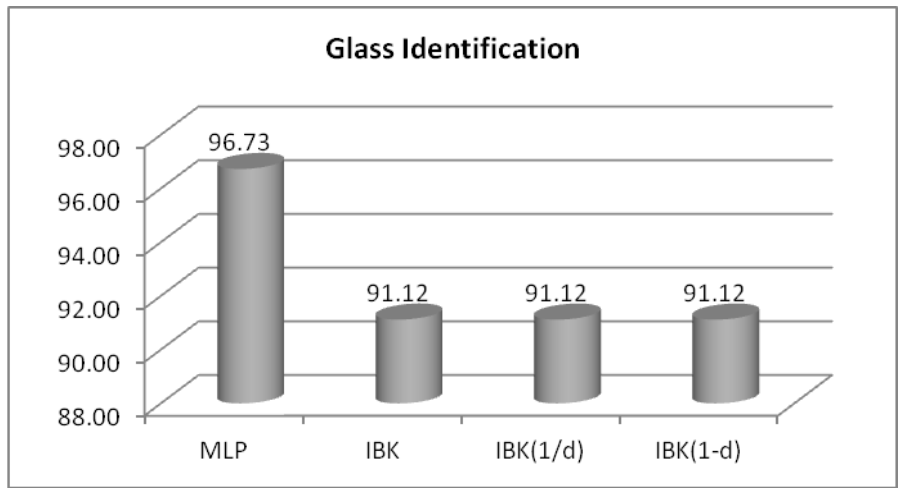

Fig. 3: Accuracy Comparison over Glass Identification

In this case MLP's performance is better than IBK and achieved 96.73% accuracy. Time consumed by IBK was 0 seconds whereas it was 1.29 seconds for MLP.

**Result:** MLP outperformed over this dataset

## 6.4  Iris

Iris has three classes for its classification problem. Fig. 4: describes the performance comparison of the two algorithms over the dataset. Where MLP took the lead over IBK, but IBK (1-distance) also achieved comparable accuracy to MLP. Since the amount of data is quiet small therefore small learning rate and momentum helps avoid over fitting and increasing accuracy.

| | |
|---|---|
| MLP: | α=0.3, β=0.3 ,Hidden Layer=4 |
| IBK: | K=9 |
| IBK(1-d): | K=9 |
| IBK(1/d): | K=9 |

Table 7: Parameter Values used for Iris

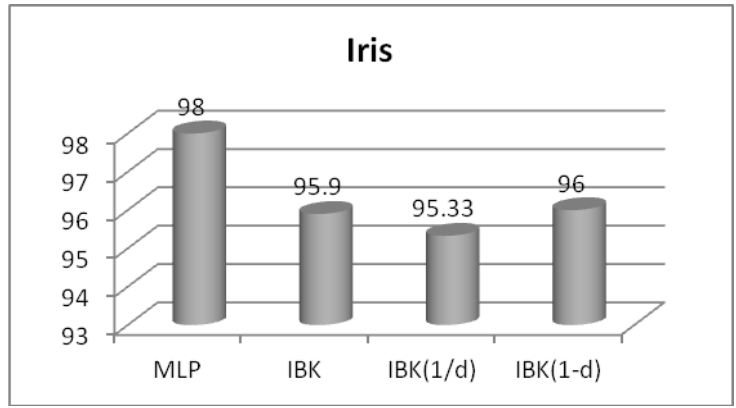

Fig. 4: Accuracy comparison over Iris

**Result:** MLP took the lead over IBK

## 6.5 Ozone Level Detection

It is a large dataset with 73 attributes, 2536 instances and 2 classes. The important thing about this dataset is like Echocardiogram it also has missing value in almost all attributes.

It is the only dataset in which all implementations of IBK performed slightly better than MLP. However, there is difference of 0.04 which is marginal as shown in the Fig. 5.

| MLP: | α=0.3 , β=0.5 ,Hidden Layer=3 |
|---|---|
| IBK: | K=11 |
| IBK(1-d): | K=11 |
| IBK(1/d): | K=11 |

Table 8: Parameter Values used for Ozone Level Prediction

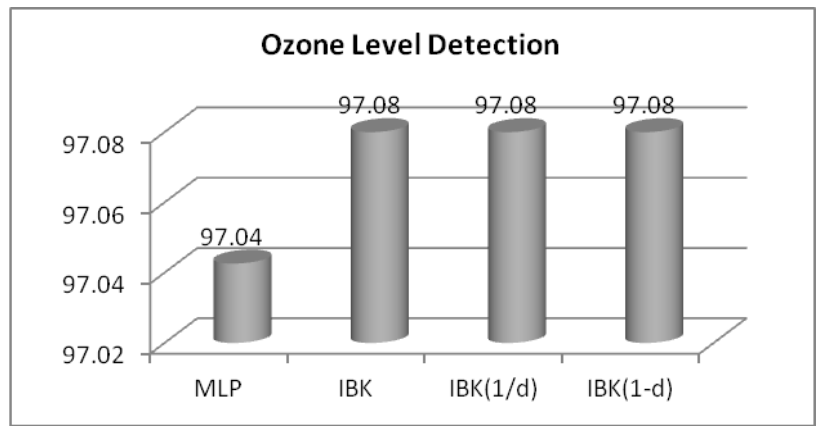

Fig. 5: Accuracy Comparison over Ozone Level Detection

Result: IBK performed slightly better.

## 6.6 Breast Cancer

The data set contains 699 instances for 10 attributes. The information contained is used to classify instances for two classes (i.e. 2 for Benign and 4 for malignant). Data set also has some missing values. 16 instances have missing data. The performance comparison shows that MLP's is a good choice for this dataset with only two hidden layers.

| MLP: α=1.0 , β=0.7 ,Hidden Layer=2 |
| --- |
| IBK: K=5 |
| IBK(1-d): K=3 |
| IBK(1/d): K=5 |

Table 9: Parameter Values used for Breast Cancer

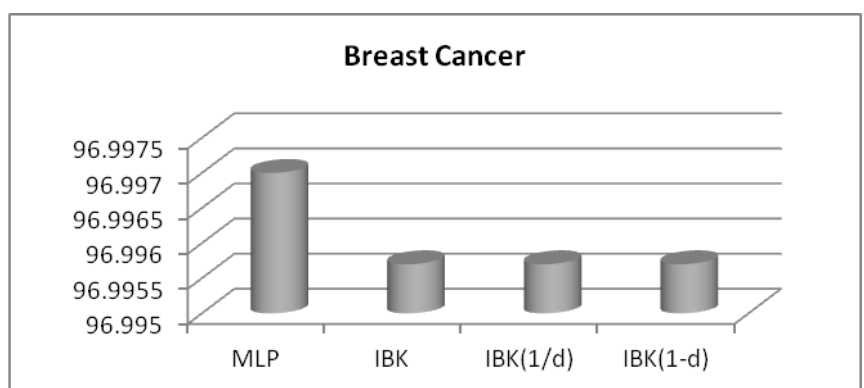

Fig. 6: Accuracy Comparison over Breast Cancer

**Result:** MLP gave better performance than IBK

## 7 Algorithm Performance Comparison

### 7.1 Root Mean Squared Value

In classification accuracy MLP took the lead over IBK with a score of 5 out of 6 in which it classified datasets better than IBK. Both algorithms work very well over linearly separable data and in case of MLP the accuracy was increased by adding hidden layers in the model and tuning its parameters (learning rate and momentum).

In order to have further comparison between two algorithms and to clearly specify which one is better some more comparing parameters has been introduced. For this purpose another important measurement of accuracy is considered i.e. Root Mean Squared Error. RMSE is the measures of the amount by which the model differ form the actual classification. The least the difference the better will be the classifier. (RMSE =0 means perfect model and 1 means worst classifier)

| S.No | Data set | MLP | IBK | IBK (1/d) | IBK (1-d) |
|---|---|---|---|---|---|
| 1 | Abalone | 0.1644 | 0.174 | 0.174 | 0.174 |
| 2 | Glass Identification | 0.0933 | 0.157 | 0.159 | 0.157 |
| 3 | Iris | 0.1233 | 0.128 | 0.130 | 0.129 |
| 4 | Echocardiogram | 0.2483 | 0.232 | 0.228 | 0.230 |
| 5 | Ozone Prediction | 0.162 | 0.161 | 0.161 | 0.161 |
| 6 | Breast Cancer | 0.1531 | 0.1566 | 0.1563 | 0.1565 |

Table 10: RMSE Comparison of Algorithms over all Datasets

Table:16 contains the Root Mean Squared values observed during the experiments. The experiment's results over datasets shown at S.No. 1-3 and 6, MLP shows smallest RMSE, especially for Glass Identification dataset the performance is outstanding and hence MLP can be said a better classifier than IBK. But in case of experiment shown at S.No 4 above, (Echocardiogram) MLP did not perform well. However, all algorithms are at par in case of experiment mentioned at S.No 5.

**Result:** MLP is better classifier

## 7.2 Kappa Statistics

"It is the measure of improvement of the model as a proportion of expectation from perfect predictor. For a predictor Kappa statistics of 1 means perfect predictor and 0 means failed" (2). In the view of above definition MLP is approaching the value of 1 or the highest value for most of the datasets and hence can be regarded better than IBK.

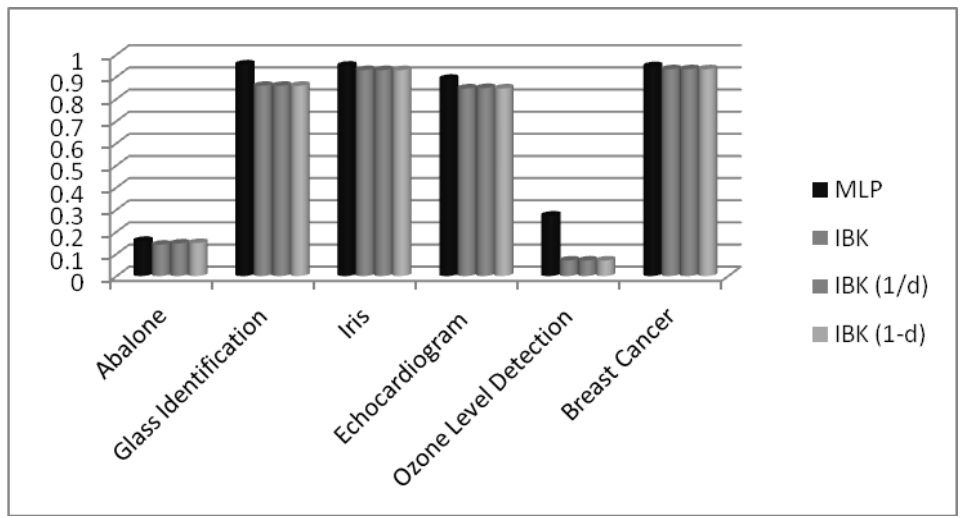

Fig. 7: Comparison based on Kappa Statistics

**Result:** MLP is preferred over IBK

## 7.3 False positive Vs True Positive Rate:

Accuracy measurement for an algorithm always take true positive result into account means the number of instances correctly classified, which does not tell how many time the algorithm predicted false value. Therefore in order to have complete analysis, comparison of both False Positive and True positive rate is important. Drawing FP rate against TP rate is known as ROC[3] curve. For a perfect predictor the ROC curve should be closer to the top left corner of the graph.

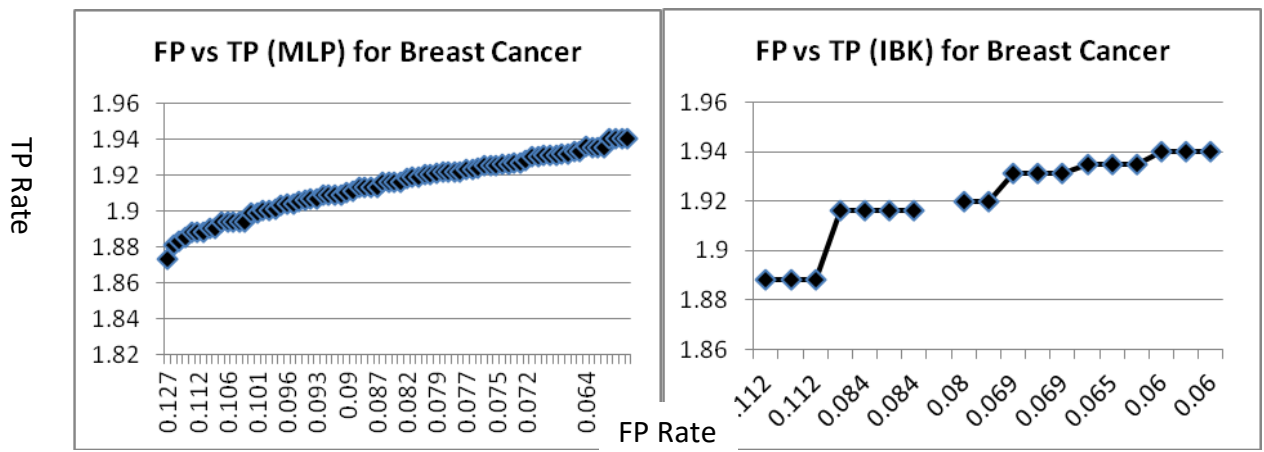

Fig. 8: ROC Curve for Breast Cancer Data set

To compare Back Propagation with Instance Based learning the curve has been drawn for Breast cancer dataset because the accuracy comparison showed very small difference. The curve shows that MLP is better classifier than IBK. As the TP rate increased smoothly in upward direction while in IBK the TP rate drops for some parameter values. It is also established that tuning MLP parameter vales increase the true positive rate and reduce the false positive rate.

# 8 Handling Missing Values

## 8.1 Introduction to Techniques

The next task is to evaluate the methods of handling missing values and comparing the performance of the two algorithms over the datasets with missing values replaced using any of the following methods.

According to the classification algorithms three methods have been explored.

First:    Default method of handling missing values implemented in WEKA. i.e.

        **MLP:** Ignore the attribute with missing values for particular instance

---

[3] Receiver Operating Characteristic curve.

**IBK:** Assign Maximum distance to the instance with missing value so that its effect over classification can be reduced.

The results using first method have been discussed in performance comparison of the two algorithms and therefore evaluation of the following two methods is given hereunder.

Second: Replacing missing values with mean/mode value of the attribute. If the attribute is real then replacement is done using mean value and if the attribute is nominal then mode (most frequently occurring) value of the attribute will replace the missing one.

Third: Missing values were replaced by using a statistical method known as Multiple Imputation.

<u>Mean/Mode Value Replacement method:</u> In this method the values are replaced according to the attribute. Mode value of the nominal attributes and Mean value of the real attribute. For this study the replacement is done using WEKA as it provides the provision of replacing missing values.

<u>Multiple Imputations</u>: "Imputation is the substitution of missing values in the datasets. Multiple Imputation (MI) is a Monte Carlo technique in which the missing values are replaced by m>1 simulated versions, each of the simulated complete datasets is analyzed by standard methods, and the results are combined to produce estimates and confidence intervals that incorporate missing-data" (1). Scheffer recommended this method to apply wherever possible instead of ignoring it. According to his recommendation MI works well if the percentage of missing values is more than 25% (3). Therefore datasets selected for comparison does have different range of missing value. There are quiet a lot of software available to replace missing values using Multiple Imputation. Here software known as NORM has been used to impute missing values in the datasets.

<u>Percentage of Missing Values in Datasets</u>

Three datasets from the above list have been used to carry out the experiments. These are Echocardiogram and Ozone Level Detection and Breast Cancer.

In first dataset the amount of data missing is almost 44% in 12 of its attributes. This also includes the class attribute.

Second dataset (Ozone Level Detection) have nearly 15% of missing data in 71 out of 73 attributes.

While Breast Cancer Dataset has only 2% of missing data in one attribute. Therefore it is valuable to address the issue of selecting replacement technique according to the percentage of data missing.

## 8.2 Performance Evaluation of algorithms:

After replacing missing values using above three methods, the datasets were then exposed to the classifiers using WEKA machine. A comparison of the results generated by classifiers is presented in the following tables.

### 8.2.1 Echocardiogram

The comparison of RMSE clearly shows that using Multiple Imputation to replace missing values reduce the root mean squared error with substantial difference and IBK algorithm classified the data with RMSE=0.1643 and kappa statistics of 0.8007. In IBK the performance is better due to the fact that the examples which could not take part in voting process now can contribute to the classification process. Hence Multiple Imputation method provides better values replacement strategy if the nature of the data is integer or real.

| Using Multiple Imputation | MLP | IBK | IBK(1/d) | IBK(1-d) |
|---|---|---|---|---|
| RMSE | 0.2817 | 0.1643 | 0.2878 | 0.2895 |
| Kappa Statistics | 0.7977 | 0.8007 | 0.8007 | 0.8007 |
| Time Taken | 0.47 | 0 | 0 | 0 |
| Using Mean/Mode | MLP | IBK | IBK(1/d) | IBK(1-d) |
| RMSE | 0.2927 | 0.2759 | 0.267 | 0.2672 |
| Kappa Statistics | 0.6842 | 0.6188 | 0.6174 | 0.6358 |
| Time Taken | 1.12 | 0 | 0 | 0 |

Table 11: Comparison of handling missing values methods for Echocardiogram

**Result:** Multiple Imputation Methods appears better than Mean/Mode value Method

### 8.2.2 Ozone Level Detection

Similar to the Echocardiogram dataset following tables shows the performance evaluation of missing value replacement techniques over Ozone dataset. The best combination of RMSE and Kappa statistics achieved is from MLP with RMSE=0.1739 and Kappa=0.3111. The Kappa statistics shows that this classification is close to perfect predictor as compared to other algorithms and this is achieved by using multiple imputation method. The time is the main factor while using MLP which is due to the use of more hidden layers to tune the model.

| Using Multiple Imputation | MLP | IBK | IBK(1/d) | IBK(1-d) |
|---|---|---|---|---|
| RMSE | 0.1739 | 0.1603 | 0.16 | 0.162 |
| Kappa Statistics | 0.3111 | 0.1341 | 0.1124 | 0.1341 |
| Time Taken | 96.24 | 0.02 | 0.02 | 0.02 |
| Using Mean/Mode | MLP | IBK | IBK(1/d) | IBK(1-d) |
| RMSE | 0.1781 | 0.1596 | 0.1595 | 0.1596 |
| Kappa Statistics | 0.2731 | 0.0703 | 0.0703 | 0.0703 |
| Time Taken | 97.69 | 0.03 | 0.04 | 0.01 |

Table 12: Comparison of Handling Missing Values methods for Echocardiogram

**Result:** Multiple Imputation Method is better than Mean/Mode replacement method

### 8.2.3 Breast Cancer

This data set was selected to analyse the impact of quantum of missing data on the technique used. Since it has only 2% of missing data, therefore the results are compared while keeping this in mind. As expected the values of RMSE and Kappa statistics suggests that Mean/Mode methods is more suitable for the data which has much less missing values.

| Using Multiple Imputation | MLP | IBK | IBK(1/d) | IBK(1-d) |
|---|---|---|---|---|
| RMSE | 0.1807 | 0.1582 | 0.1585 | 0.1581 |
| Kappa Statistics | 0.9247 | 0.9307 | 0.9307 | 0.9307 |
| Time Taken | 2 | 0 | 0 | 0.01 |
| Using Mean/Mode | MLP | IBK | IBK(1/d) | IBK(1-d) |
| RMSE | 0.1696 | 0.1565 | 0.1562 | 0.1564 |
| Kappa Statistics | 0.9373 | 0.9307 | 0.9337 | 0.9307 |
| Time Taken | 2 | 0 | 0 | |

Table 13: Comparison of Handling Missing Values methods for Breast Cancer

**Result:** Mean/Mode replacement method is well suited for this dataset

## 9 Conclusion

The back propagation algorithm used to adjust weights in Multi Layer Perceptron Model appears to be better technique for classification problems as compared to Instance Based learning method. It has been observed that achieving best parameter combination for MLP is a difficult task but usually produce impressive results. While Instance Based learning cannot be classified as a bad option because it has very simple implementation and needs negligible time to classify with comparable accuracy with back propagation. The Root mean Square, Kappa statistics and ROC curve proves Back propagation to be more consistent to get good accuracy with low error rates. The lower values of Learning rate, momentum and hidden layers also helped to avoid over fitting which in turn pushed Back Propagation to have better Kappa Statistics. Therefore it would be fair to say that if the data is very sensitive and the emphasis is on the reduced error rate then Back Propagation should be preferred over Instance Based Learning.

In case of handling Missing values, Multiple Imputation is better in handling the problem but the study suggests that this method should be used if the amount of missing data is roughly more than 10%. It is evident from the experiments over Breast Cancer Dataset that below this percentage mean/mode replacement methods give better results. But it needs a lot more experiments to have confidence over this percentage of missing data.